\documentclass[journal,twoside,web]{ieeecolor}
\usepackage{generic}
\providecommand{\refname}{References}
\usepackage{stfloats}
\usepackage{booktabs}
\usepackage{cite}
\usepackage{booktabs}
\usepackage{colortbl}
\usepackage{xcolor}
\usepackage{graphicx}
\definecolor{mygray}{gray}{0.88}
\newcommand{\thickhline}{\noalign{\global\arrayrulewidth=1pt}\hline\noalign{\global\arrayrulewidth=0.4pt}}
\usepackage{amsmath,amssymb,amsfonts}
\usepackage{algorithmic}
\usepackage{graphicx}
\usepackage{algorithm,algorithmic}
\usepackage{hyperref}
\hypersetup{hidelinks=true}
\usepackage{textcomp}
\def\BibTeX{{\rm B\kern-.05em{\sc i\kern-.025em b}\kern-.08em
    T\kern-.1667em\lower.7ex\hbox{E}\kern-.125emX}}
\begin{document}

\title{Understanding From Human Perspective: A Multi-agent System for Interactive Egocentric Medical Image Segmentation}

\author{Rongjun Ge, Dongyang Wang, Heng Zhu, Zhirui Li, Yang Chen, and Yuting He
\thanks{This work was supported by the National Natural Science Foundation of China under Grant T2225025.}
\thanks{Rongjun Ge and Heng Zhu are with the School of Instrument Science and Engineering, Southeast University, Nanjing 210096, China (e-mail: rongjun\_ge@seu.edu.cn; zhumoran519@gmail.com).}
\thanks{Dongyang Wang and Yang Chen are with the School of Computer Science and Engineering, Southeast University, Nanjing 210096, China (e-mail: 220255090@seu.edu.cn; chenyang.list@seu.edu.cn).}
\thanks{Zhirui Li is with the Sichuan University-Pittsburgh Institute, Sichuan University, Chengdu 610065, China (e-mail: lizhirui@stu.scu.edu.cn).}
\thanks{Yuting He is with the Department of Biomedical Engineering, Case Western Reserve University, Cleveland, OH 44106 USA (e-mail: yuting.he4@case.edu).}
\thanks{(Corresponding authors: Yang Chen and Yuting He.)}}

\maketitle

\begin{abstract}
Interactive egocentric medical image segmentation (IEMIS) plays an important role in smart-glasses-assisted medical image review, segmenting the medical targets a clinician refers to from their egocentric view. Once it succeeds, the object-level visual evidence it provides strengthens the review and underpins fine-grained analysis and clinical decision-making. However, the instruction and the video both come from the user's egocentric perspective, which poses two challenges. (1) Semantic ambiguity leaves the model unable to confirm the user-intended target. (2) Visual variability makes the segmentation jump from frame to frame. In this paper, we propose EgoMed-Agent, a multi-agent system that understands the target from the human perspective through two workflows. (1) The \textit{Target Confirmation Workflow} grounds the instruction against candidate targets with a reliability score, confirming the target when the grounding is reliable and asking the user to clarify when it is not, thereby confirming the segmentation target. (2) The \textit{Localization-Guided Propagation Workflow} couples mask propagation with per-frame target localization, using the localized target to correct the propagated mask whenever the two diverge, so the segmentation stays on the target across the egocentric video. Extensive experiments show that EgoMed-Agent reaches 71.34\% average Dice, far above the best text-prompted baseline (11.70\%). Our code is available at \href{https://github.com/wdyyyyyy/EgoMed-Agent}{our project page}.

\end{abstract}

\begin{IEEEkeywords}
Interactive egocentric medical image segmentation, multi-agent system, medical smart glasses, egocentric vision.
\end{IEEEkeywords}

\section{Introduction}
\label{sec:introduction}

\begin{figure}[t]
  \centering
  \includegraphics[width=\columnwidth]{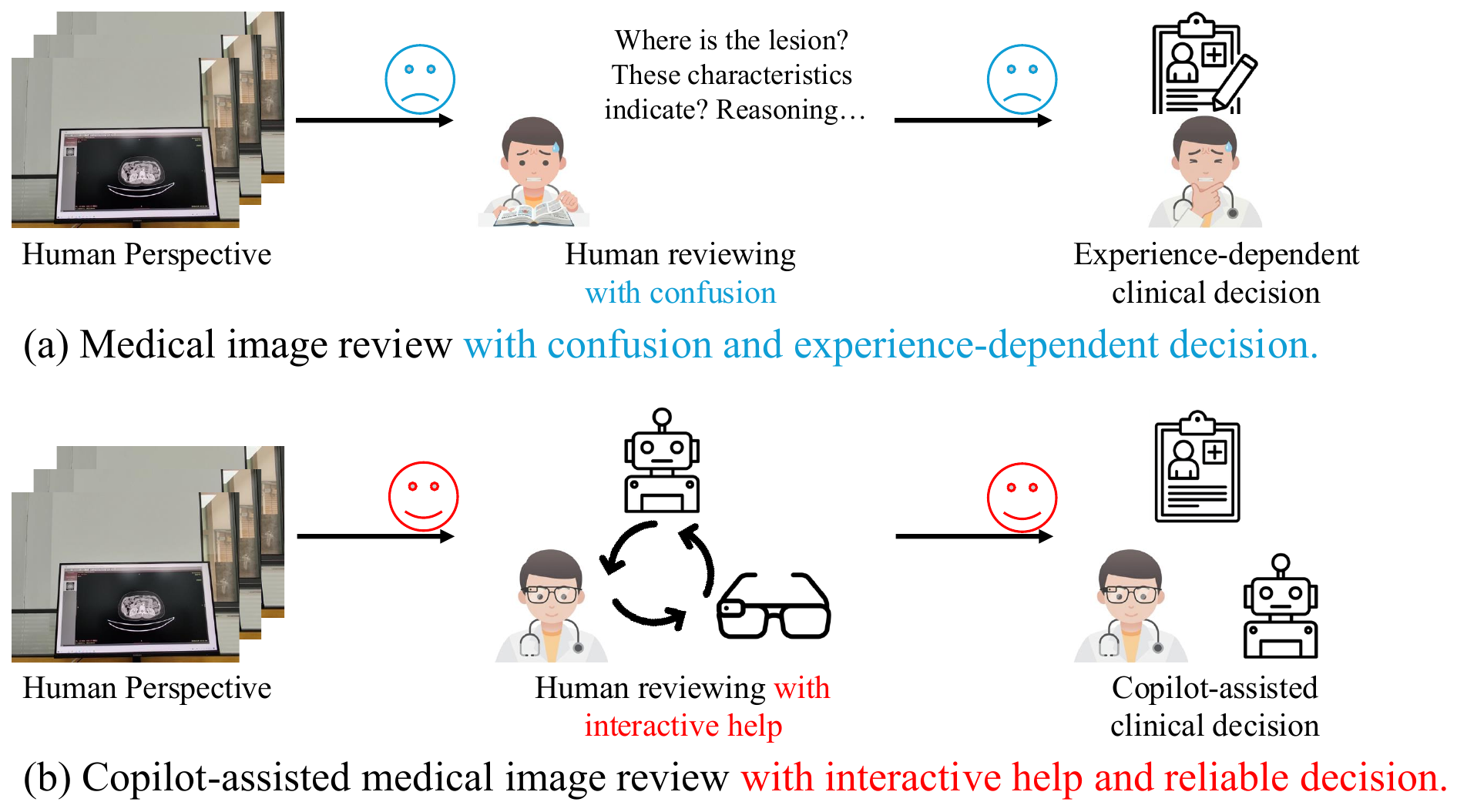}
  \caption{(a) Medical image review leaves the doctor without support, so the clinical decision is experience-dependent. (b) Smart-glasses-assisted medical image review gives the doctor support, so the clinical decision is reliable.}
  \vspace{-10pt}
  \label{fig:task}
\end{figure}

\begin{figure}[t]
  \centering
  \includegraphics[width=\columnwidth]{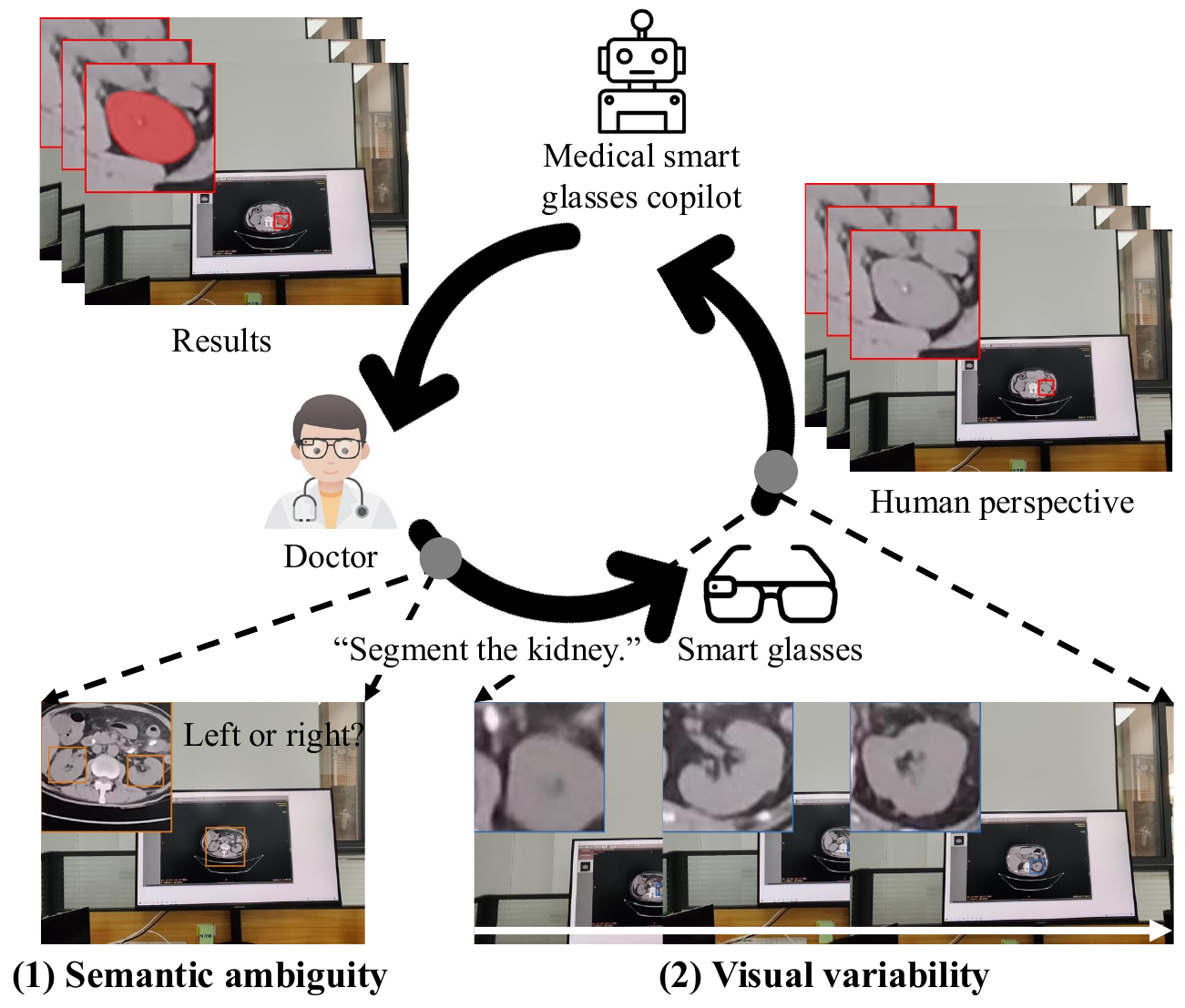}
  \caption{IEMIS is formulated as an interactive loop among the doctor, smart glasses, and the medical smart glasses copilot. In each interaction cycle, the doctor issues an instruction, the smart glasses capture the egocentric video stream and forward it to the copilot, and the copilot returns the corresponding segmentation results (top). This egocentric interaction poses two challenges (bottom). (1) \textit{Semantic ambiguity.} The instruction ``segment the kidney'' matches both the left and right kidney, leaving the model unable to determine the user-intended target. (2) \textit{Visual variability.} Head motion, viewpoint changes, and lighting variations make the target's position, scale, and shape vary across frames, so its segmentation jumps from frame to frame.}
  \vspace{-10pt}
  \label{fig:iemis}
\end{figure}

\IEEEPARstart{I}{nteractive} egocentric medical image segmentation (IEMIS) plays an important role in medical smart glasses copilots~\cite{malungana2025use,zhang2022designing}.
Medical image review is a continuous reasoning process in which doctors identify lesions, inspect anatomical structures, and integrate visual evidence into diagnostic decisions~\cite{waite2017interpretive}.
During this process, the reviewer faces confusion over where a lesion lies and what its characteristics indicate, relying on their own expertise~\cite{yu2014radiologist}, so clinical decisions become experience-dependent and vulnerable when critical findings are overlooked~\cite{drew2013invisible} (Fig.~\ref{fig:task}(a)).
Medical smart glasses copilots have achieved preliminary success in clinical settings~\cite{malungana2025use,zhang2022designing}.
By perceiving the review context from the doctor's egocentric perspective, these copilots enable an interactive paradigm in which the doctor issues instructions during image review while the copilot answers queries and provides suggestions for clinical decision-making (Fig.~\ref{fig:task}(b)).
In this paradigm, IEMIS provides the perceptual basis: it segments instruction-relevant medical objects from egocentric video streams (Fig.~\ref{fig:iemis}).
Once IEMIS succeeds, the copilot grounds its answers on concrete object-level visual evidence, such as lesions and anatomical structures, providing the perceptual foundation for fine-grained analysis and evidence-based clinical decision support.

However, both the instruction and the video in IEMIS come from the user's egocentric perspective, which poses two challenges (Fig.~\ref{fig:iemis}).
(1) \textit{Semantic ambiguity in interactions.} User instructions exhibit semantic ambiguity~\cite{su2025ova,zhang2024clamber,ding2026open}.
For example, the instruction ``Segment the kidney.'' does not specify whether the left or right kidney is intended, leaving both as plausible candidate targets. Such ambiguity leaves the model unable to determine the user-intended target from the instruction alone.
(2) \textit{Visual variability from egocentric perspectives.} Egocentric video streams contain substantial visual variability caused by head motion, viewpoint changes, and lighting variations~\cite{li2026challenges,poleg2014temporal,millerdurai2024eventego3d}. These variations cause the target to appear with changing position, scale, and shape across consecutive frames, so its segmentation jumps from frame to frame~\cite{wang2025caldiff}.

Multi-agent systems (MAS) tackle complex tasks by decomposing them into coordinated subtasks assigned to agents with complementary capabilities~\cite{li2024agent,zhang2025agentorchestra,hu2026evolutionary}.
Decomposing IEMIS along these two challenges lowers the interaction complexity of confirming and segmenting the user-intended target, which makes MAS a suitable paradigm for IEMIS.
(1) For semantic ambiguity, MAS decomposes target confirmation into candidate perception and reliability-based grounding. Different agents identify possible medical targets and determine the user-intended one from the instruction, confirming what to segment before segmentation begins.
(2) For visual variability, MAS decomposes the video segmentation into temporal propagation and target-aware localization. Different agents maintain the user-intended target across frames and localize it in each frame, keeping the segmentation on the user-intended target across the egocentric video.

In this paper, we propose the Interactive \textbf{Ego}centric \textbf{Med}ical Image Segmentation \textbf{Agent} (EgoMed-Agent), the first multi-agent system for IEMIS. EgoMed-Agent decomposes IEMIS into two workflows implemented by three specialized agents, a Detection Agent, a Confirmation Agent, and a Propagation Agent.
(1) The Target Confirmation Workflow confirms the segmentation target, asking the user to clarify only when the instruction does not reliably identify the target. The Confirmation Agent grounds the instruction against the candidate targets from the Detection Agent and scores the reliability of this grounding. A reliable grounding directly fixes the user-intended target, while an unreliable one triggers a clarification that completes the user's intent. Through this reliability-based grounding and clarification, the workflow determines the user-intended target, so the subsequent segmentation acts on an explicit target.
(2) The Localization-Guided Propagation Workflow couples mask propagation with per-frame target localization, using the localized target to correct the propagated mask whenever the two diverge. The Propagation Agent segments the user-intended target from its localized position and propagates the mask for temporal continuity, while the Detection Agent re-localizes the target in each frame. The system evaluates the consistency between the propagated result and this localization and re-initializes propagation from the detected position whenever they diverge, so the segmentation stays on the target across the egocentric video.

Specifically, our contributions are summarized as follows:
\begin{itemize}
    \item We propose IEMIS, which segments instruction-relevant medical objects from the human perspective to strengthen smart-glasses-assisted medical image review, which in turn underpins precise localization of structures, fine-grained analysis, and clinical decision support.

   \item We propose EgoMed-Agent, the first multi-agent system for IEMIS, which decomposes the task into the Target Confirmation and Localization-Guided Propagation workflows, so the system confirms the user-intended target before segmentation and keeps the segmentation on that target across the egocentric video.

    \item We propose the Target Confirmation Workflow, in which the Detection Agent perceives candidate targets and the Confirmation Agent grounds the instruction against them with a reliability score, requesting clarification from the user only when the grounding is unreliable, thereby determining the user-intended target before segmentation.

    \item We propose the Localization-Guided Propagation Workflow, in which the Detection and Propagation Agents couple mask propagation with per-frame target localization. The localized target corrects the propagated mask whenever the two diverge, keeping the segmentation on the user-intended target across the egocentric video.

    \item Extensive experiments show that EgoMed-Agent reaches $71.34\%$ average Dice and significantly outperforms the strongest text-prompted baseline ($11.70\%$, $p<0.001$ on every target), with strong potential for smart-glasses-assisted medical image review.
\end{itemize}

Overall, our EgoMed-Agent has three key advantages: a) \textit{Reliable target understanding:} By confirming the user-intended target from the human perspective before segmentation, EgoMed-Agent grounds the segmentation on the target the clinician refers to, bringing reliable instruction-guided segmentation on egocentric medical scenes. b) \textit{Stable cross-frame segmentation:} By coupling mask propagation with per-frame target localization, EgoMed-Agent corrects propagation drift under head motion and viewpoint change, keeping the segmentation on the user-intended target across the egocentric video. c) \textit{Training-free generality:} As a general multi-agent paradigm that assembles off-the-shelf detection, grounding, and propagation models without task-specific training, EgoMed-Agent applies across diverse medical modalities and real-world capture scenes.

\section{Related Work}

\subsubsection{Egocentric Vision}

Egocentric vision studies visual understanding from the user's egocentric perspective, providing a natural perception paradigm for wearable devices and human-centered assistance~\cite{li2026challenges}.
Large-scale egocentric benchmarks, such as EPIC-KITCHENS~\cite{damen2018scaling} and Ego4D~\cite{grauman2022ego4d}, have further advanced egocentric video understanding.
Multimodal large language models are also evaluated on egocentric video understanding, including fine-grained temporal reasoning and scene-text question answering~\cite{plizzari2025omnia,zhou2025egotextvqa}.
More recent works segment objects from egocentric video through language, via egocentric referring video object segmentation~\cite{liu2025ceres} and text-supervised egocentric semantic segmentation~\cite{shi2024cognition}.
In the medical domain, egocentric vision has been used to interpret clinical scenes from egocentric videos, including surgical workflows, medical actions, and instrument usage~\cite{fujii2024egosurgery,zhuo2025egocentric,darjana2025egosurgery}.
However, these studies either interpret the surrounding clinical environment or segment general daily objects by mapping language directly to masks.
In contrast, IEMIS focuses on segmenting user-intended medical targets from egocentric video streams captured during interaction with medical images.
This involves identifying the user-intended target from user instructions and keeping the segmentation aligned with that target across frames, which existing formulations do not support.

\subsubsection{Copilots}

AI copilots are interactive assistant systems that perceive user context, respond to instructions, and provide task-oriented support~\cite{plizzari2024outlook}. Smart glasses extend copilots to wearable and hands-free scenarios, enabling them to operate from the user's egocentric perspective~\cite{yang2026egocentric}. In the medical domain, smart-glasses copilots have been explored for workflow assistance, such as remote consultation, surgical training, and clinical documentation~\cite{gollapalli2024smart,malungana2025use,zhang2022designing,mitrasinovic2015clinical}, and generative copilots support clinicians in medical image interpretation and question answering~\cite{lu2024multimodal}. However, IEMIS centers on object-level visual grounding, where the copilot localizes and segments the medical target specified by the user from the egocentric video streams. Such object-level perception remains underexplored in current medical copilot systems.

\subsubsection{Interactive Medical Image Segmentation}

Interactive medical image segmentation allows users to delineate medical objects through prompts such as points, boxes, scribbles, and texts~\cite{xu2024advances,ma2024segment,wang2018deepigeos}.
Among these prompts, text offers an intuitive interaction modality in egocentric settings, allowing users to specify target medical objects naturally through language.
Text-prompted segmentation therefore provides a relevant formulation for hands-free medical image segmentation.
These methods formulate the task as segmenting the object described by a given textual prompt~\cite{heinemann2025limis,yuan2025tgsam,liu2025medsam3, lai2024lisa}, making the target depend on the prompt description. 
More recent efforts adapt promptable segmentation foundation models to medical images, producing segmentation from visual or textual prompts~\cite{li2024prism,xie2024simtxtseg,shen2023temporally}.
However, real interactions involve incomplete instructions that omit discriminative target details, leading to semantic ambiguity.
This motivates confirming the target before performing segmentation in EgoMed-Agent.

\subsubsection{Multi-agent Systems}

Multi-agent systems (MAS) coordinate specialized agents to tackle tasks beyond the reach of a single model, and have been applied across general and medical domains~\cite{li2024agent,zhang2025agentorchestra,hu2026evolutionary}.
In particular, existing medical MAS have been developed for clinical diagnosis simulation, radiology report generation and evaluation, and tool-augmented clinical reasoning~\cite{schmidgall2024agentclinic,almansoori2025medagentsim,elboardy2025medical,yi2025multimodal,liao2025reflectool}.
Unlike these, IEMIS targets a different MAS setting, where agents interactively determine the user-intended visual target from instructions and keep the segmentation aligned with that target across the egocentric video.
This couples instruction-guided target confirmation with object-level visual perception and video segmentation.

\section{Methodology}

\begin{figure*}[t]
  \centering
  \includegraphics[width=\textwidth]{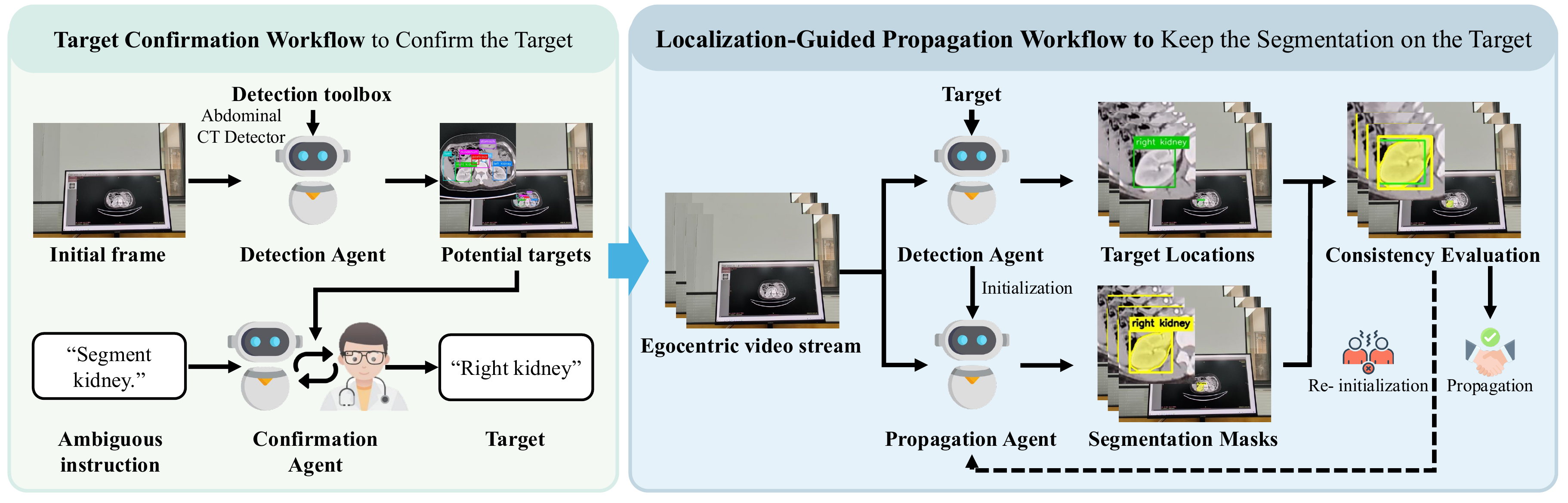}
  \caption{EgoMed-Agent decomposes IEMIS into two cooperating agent workflows. The Target Confirmation Workflow, run by the Detection and Confirmation Agents, confirms the user-intended target, while the Localization-Guided Propagation Workflow, run by the Detection and Propagation Agents, keeps the segmentation on that target across the egocentric video through Consistency Evaluation.}
  \label{fig:egomed}
\end{figure*}

Our EgoMed-Agent (Fig.~\ref{fig:egomed}) decomposes IEMIS into two workflows implemented by three specialized agents. It identifies the user-intended target from instructions through Target Confirmation (Detection and Confirmation Agents, Sec.~\ref{subsec:tc}), and keeps the segmentation on that target across the egocentric video through Localization-Guided Propagation (Detection and Propagation Agents, Sec.~\ref{subsec:lgp}). Together, the two workflows produce reliable instruction-guided medical segmentation from the human perspective.

\subsection{Formulation of Problem and Overall Framework}

\subsubsection{The Formulation of IEMIS}

IEMIS is formulated as an interactive loop among the doctor, smart glasses, and copilot.
The doctor issues instructions at different time points, and each instruction defines an interaction cycle.
In the $k$-th interaction cycle, the doctor issues an instruction $I^{k}$, the smart glasses capture the corresponding egocentric video stream $\mathcal{V}^{k}_{1:t_k}=\{v^{k}_1,v^{k}_2,\dots,v^{k}_{t_k}\}$, and the copilot predicts the mask sequence $\mathcal{M}^{k}_{1:t_k}=\{m^{k}_1,m^{k}_2,\dots,m^{k}_{t_k}\}$:
\begin{equation}
    \mathcal{M}^{k}_{1:t_k} = \mathcal{F}(\mathcal{V}^{k}_{1:t_k}, I^{k}),
\end{equation}
where $m^{k}_i$ denotes the mask of the user-intended medical object in frame $v^{k}_i$, and $\mathcal{F}$ denotes the IEMIS model.
The index $k$ distinguishes different interaction cycles within a continuous egocentric video stream.
For simplicity, we omit the superscript $k$ in the following sections and denote one interaction cycle as $(I, \mathcal{V}_{1:t}, \mathcal{M}_{1:t})$.

\subsubsection{The Framework of EgoMed-Agent}
EgoMed-Agent decomposes IEMIS into two workflows supported by three specialized agents, a Detection Agent $A_{\text{det}}$, a Confirmation Agent $A_{\text{con}}$, and a Propagation Agent $A_{\text{pro}}$.

In the Target Confirmation Workflow, the Detection Agent $A_{\text{det}}$ and the Confirmation Agent $A_{\text{con}}$ collaborate to determine the user-intended target $\theta$ from the initial frame $v_1$ and instruction $I$. This workflow confirms what should be segmented.
In the Localization-Guided Propagation Workflow, the Detection Agent $A_{\text{det}}$ and the Propagation Agent $A_{\text{pro}}$ collaborate to generate the mask sequence $\mathcal{M}_{1:t}$ for the user-intended target $\theta$ across the egocentric video $\mathcal{V}_{1:t}$. This workflow keeps the segmentation on the user-intended target throughout the video stream. The overall process is formulated as:
\begin{equation}
    \theta = \mathcal{W}_1(v_1, I;\, A_{\text{det}}, A_{\text{con}}), \quad
    \mathcal{M}_{1:t} = \mathcal{W}_2(\mathcal{V}_{1:t}, \theta;\, A_{\text{det}}, A_{\text{pro}}),
\end{equation}
where $\mathcal{W}_1$ and $\mathcal{W}_2$ denote the Target Confirmation and Localization-Guided Propagation workflows, each carried out by the agents listed after the semicolon.

\subsection{Target Confirmation Workflow to Confirm the Target}
\label{subsec:tc}

Target Confirmation determines the user-intended segmentation target $\theta$ before segmentation.
It is implemented by the Detection Agent $A_{\text{det}}$ and the Confirmation Agent $A_{\text{con}}$.
The Detection Agent extracts candidate medical targets from the initial frame, and the Confirmation Agent grounds the user instruction to these candidates. Through modality-specific candidate extraction and reliability-based grounding with clarification, this workflow confirms the user-intended target, so the subsequent segmentation acts on an explicit target.

\subsubsection{Detection Agent}

The Detection Agent extracts potential medical targets from the initial frame.
To support different medical image types, it is equipped with a toolbox $\mathcal{T}$ containing a set of specialized detectors.
Given the initial frame $v_1$, the Detection Agent first predicts the imaging type $c$, selects the corresponding detector $d$ from the toolbox, and then applies the selected detector to obtain the candidate target set $O$:
\begin{equation}
    c = C_{\text{mod}}(v_1), \quad d = \mathcal{T}(c), \quad O = d(v_1),
\end{equation}
where $C_{\text{mod}}$ denotes the type classifier, $d$ denotes the detector selected from $\mathcal{T}$, and $O = \{o^{(1)}, o^{(2)}, \dots, o^{(N)}\}$ denotes the set of candidate targets.
Each candidate $o^{(i)}$ contains an object category and a bounding box.
This toolbox design allows the Detection Agent to extract candidate targets across different medical imaging contexts.

\subsubsection{Confirmation Agent}

The Confirmation Agent determines the user-intended target according to the user instruction and the candidate targets extracted by the Detection Agent. Given the instruction $I$ and the candidate set $O$, the Confirmation Agent grounds the instruction to the candidates and outputs a target result $\theta^*$ together with a reliability score $s$:
\begin{equation}
    \theta^*, s = A_{\text{con}}(I, O),
    \label{eq:confidence}
\end{equation}
where $\theta^*$ denotes the target result obtained from the current grounding process, and $s$ measures the reliability of this grounding result.

Given a grounding reliability threshold $\tau_1$, if $s \geq \tau_1$, the grounding result is considered reliable, and $\theta^*$ is directly confirmed as the user-intended target $\theta$ for subsequent segmentation. If $s < \tau_1$, the grounding result is considered unreliable, indicating that the current instruction and candidates do not reliably determine the user-intended target. In this case, the Confirmation Agent proactively requests clarification from the user and updates the grounding result according to the user's clarification response $r$:
\begin{equation}
    \theta^r = A_{\text{con}}(I, O, r),
\end{equation}
where $\theta^r$ denotes the clarified target result.

The final user-intended target $\theta$ is defined as:
\begin{equation}
    \theta =
    \begin{cases}
        \theta^*, \quad \mathrm{if}\ s \geq \tau_1, \\
        \theta^r, \quad \mathrm{if}\ s < \tau_1.
    \end{cases}
    \label{eq:confirm}
\end{equation}
Through this reliability-based grounding and clarification mechanism, EgoMed-Agent confirms the user-intended target before segmentation, enabling subsequent segmentation to be conditioned on explicit user intent.

\subsubsection{Properties of Target Confirmation}

The Target Confirmation workflow provides two properties for IEMIS:
(1) \textbf{Modality-specific Candidate Perception.} By selecting specialized detectors according to the medical image type, this workflow extracts candidate targets that are specific to the current imaging context, providing reliable visual candidates for instruction grounding.
(2) \textbf{Clarification-based Target Confirmation.} By evaluating grounding reliability, this workflow identifies unreliable grounding results and triggers clarification in such cases, enabling the user-intended target to be confirmed before segmentation.

\subsection{Localization-Guided Propagation Workflow to Keep the Segmentation on the Target}
\label{subsec:lgp}

Localization-Guided Propagation generates segmentation masks for the target across the egocentric video $\mathcal{V}_{1:t}$ through the collaboration of the Detection Agent $A_{\text{det}}$ and the Propagation Agent $A_{\text{pro}}$. The Detection Agent provides target-aware localization results in video frames, while the Propagation Agent segments the target according to the localized position and propagates its mask over time. By combining target localization with temporal propagation, this workflow keeps the segmentation on the target across the egocentric video.

\subsubsection{Propagation Agent}

The Propagation Agent maintains the segmentation mask of the user-intended target $\theta$ across the egocentric video. Specifically, after target confirmation, the Detection Agent localizes $\theta$ in the initial frame $v_1$ using the specialized detector selected in the Target Confirmation Workflow and provides the corresponding bounding box $b^{\text{loc}}_1$. This bounding box is used to initialize the Propagation Agent. Starting from this initialization, the Propagation Agent generates the initial mask and propagates it through subsequent frames to produce a temporally continuous mask sequence:
\begin{equation}
\mathcal{M}_{1:t} = A_{\text{pro}}(\mathcal{V}_{1:t}, b^{\text{loc}}_1).
\end{equation}
Through this localized initialization and temporal propagation, the Propagation Agent produces a temporally continuous mask sequence for the user-intended target.

\subsubsection{Consistency Evaluation}

Consistency Evaluation keeps the propagated masks on the user-intended target across the egocentric video by correcting propagation drift with target-aware localization. For each frame $v_i$, the Detection Agent uses the specialized detector selected in the Target Confirmation Workflow to localize the user-intended target with a bounding box $b^{\text{loc}}_i$. Meanwhile, the propagated mask produced by the Propagation Agent is converted into a propagated bounding box $b^{\text{pro}}_i$ for Consistency Evaluation. The consistency between the localized target and the propagated result is measured by their IoU, $\mathrm{IoU}_i = |b^{\text{loc}}_i \cap b^{\text{pro}}_i| / |b^{\text{loc}}_i \cup b^{\text{pro}}_i|$.

Given a consistency threshold $\tau_2$, when $\mathrm{IoU}_i \geq \tau_2$, the propagated result is considered consistent with the localized target, and the Propagation Agent continues propagation. When $\mathrm{IoU}_i < \tau_2$, the propagated result is considered unreliable. In this case, the system re-initializes the Propagation Agent using the localized bounding box $b^{\text{loc}}_i$ determined by the Detection Agent, bringing the propagation process back to the target. Through this mechanism, EgoMed-Agent preserves temporal continuity when the propagated result is reliable and uses target-aware localization to re-initialize propagation when inconsistency occurs. This keeps the segmentation on the user-intended target across the egocentric video.
\subsubsection{Properties of Localization-Guided Propagation}

The Localization-Guided Propagation workflow provides two properties for IEMIS:
(1) \textbf{General Propagation Capability.} The Propagation Agent propagates masks in a target-agnostic manner after initialization. Therefore, once the Detection Agent provides the initial target location, the propagation mechanism applies to different medical targets.
(2) \textbf{Target-Aware Re-localization.} By evaluating the consistency between the propagated result and the Detection Agent localization, the system determines whether propagation remains aligned with the user-intended target. When inconsistency occurs, the target-aware localization result is used to re-initialize propagation, correcting propagation drift to keep the segmentation on the user-intended target across the egocentric video.

\section{Experiments}

\subsection{Experimental Settings}

\subsubsection{Dataset}

\begin{figure}[t]
\centering
\includegraphics[width=\linewidth]{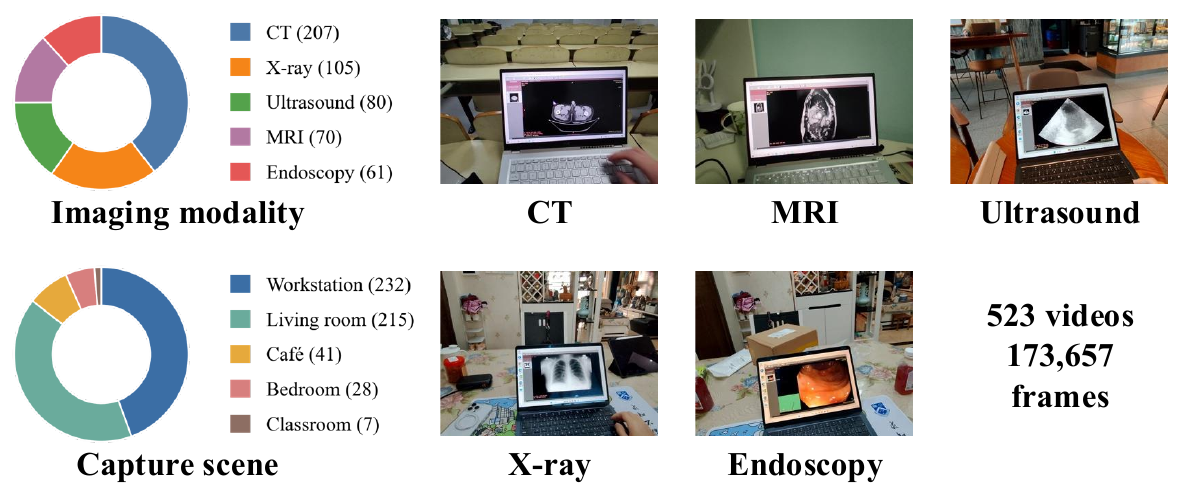}
\caption{Our dataset covers five imaging modalities and five capture scenes, recorded as egocentric video of on-screen medical images. The two pies give the number of videos per modality and per capture scene, and the panels on the right show one egocentric example per modality.}
\label{fig:dataset_dist}
\end{figure}

We construct an egocentric medical image segmentation dataset for IEMIS, containing 523 videos and 173,657 frames that span 5 imaging modalities, 12 medical targets, and 5 real-world capture scenes, as shown in Fig.~\ref{fig:dataset_dist}. The videos are collected from five public medical image sources, spanning CT (AMOS~\cite{ji2022amos}), MRI (ACDC~\cite{bernard2018deep}), ultrasound (CAMUS~\cite{leclerc2019deep}), X-ray (MCC~\cite{jaeger2014two}), and endoscopy (PMC~\cite{ali2023multi}). To build the egocentric reviewing scenario, we display these medical images in a DICOM viewer (RadiAnt) on a screen and record the reviewing process from the egocentric perspective with smart glasses (Xiaomi AI Glasses and Rokid Glasses), keeping the screen within view. The recordings span five everyday environments (workstation, living room, cafe, bedroom, and classroom) and cover both static viewing and head motion (left-right or forward-backward), reflecting diverse real-world reviewing conditions. For each video, we annotate the segmentation mask of every medical target and design multiple sequential instructions. The dataset is publicly available at \href{https://huggingface.co/datasets/daizywang/EgoMed-IEMIS}{HuggingFace}.

\subsubsection{Implementation Details}

We instantiate the image-type classifier in the Detection Agent using YOLO26-classify to identify the medical image type. We instantiate the five detection tools in the Detection Agent toolbox using YOLO26-det, corresponding to CT, MRI, ultrasound, X-ray, and endoscopy images. The image-type classifier and all modality-specific detectors are trained on our data using an NVIDIA RTX A6000 GPU. The dataset is split into training, validation, and test sets with a ratio of 5:2:3. On the test set, the image-type classifier reaches $100\%$ routing accuracy ($159/159$ cases), so each frame is dispatched to the correct modality-specific detector.  The Confirmation Agent is instantiated with DeepSeek-V4-Flash with
prompts to output the reliability score and user-intended target. The grounding reliability threshold in Target Confirmation is set to $\tau_1=0.8$. We instantiate the Propagation Agent using SAM2 with its original weights to perform mask segmentation and propagation. Based on the hyperparameter experiments in Sec.~\ref{ct}, we set the consistency threshold in Consistency Evaluation to $\tau_2=0.6$.

\subsubsection{Comparison Methods}

As shown in Table~\ref{tab:main_results}, we compare EgoMed-Agent with five text-prompted segmentation baselines, including Grounded SAM2~\cite{ravi2024sam2segmentimages, liu2023grounding,ren2024grounding,ren2024grounded,kirillov2023segany}, LangSAM~\cite{langsam}, LISA~\cite{lai2024lisa}, MedSAM3~\cite{liu2025medsam3}, and SAM3~\cite{carion2025sam}. These methods take the interaction instruction and the egocentric video as input and output the mask sequence of the instructed target.

We also report nnU-Net~\cite{isensee2021nnu} as a supervised upper-bound reference. Unlike instruction-guided methods, nnU-Net performs predefined semantic segmentation with fixed target categories and is trained using the mask annotations in the training split. Therefore, nnU-Net is not used as a direct comparison method, but as a reference showing the performance achievable under fully supervised category-specific segmentation.

\begin{table*}[t]
\centering
\caption{Quantitative comparison of segmentation performance, reported as target-wise Dice (\%, mean $\pm$ std over cases). For each target, the best instruction-guided result is in \textbf{bold} and the second best is \underline{underlined}. nnU-Net (gray) is a fully supervised upper-bound reference, not a competitor, and ``-'' marks a method that produced no valid mask for that target. The Average row reports the mean $\pm$ std across the twelve targets, and the last column gives the $p$-value of a paired $t$-test between EgoMed-Agent and the strongest baseline MedSAM3 on per-case Dice.}
\label{tab:main_results}
\renewcommand{\arraystretch}{1.0}
\resizebox{\textwidth}{!}{
\begin{tabular}{l|>{\color{gray}}c|ccccc|c|c}
\thickhline
\textbf{Targets} & \textbf{nnU-Net} & \textbf{Grounded SAM2} & \textbf{LangSAM} & \textbf{LISA} & \textbf{MedSAM3} & \textbf{SAM3} & \textbf{EgoMed-Agent} & \textbf{$p$-value} \\
\hline
AMOS liver & $82.82 \pm 13.9$ & $\underline{19.80 \pm 11.2}$ & $17.89 \pm 11.9$ & $11.52 \pm 9.7$ & $15.15 \pm 21.7$ & $0.03 \pm 0.2$ & $\mathbf{71.69 \pm 15.6}$ & $3.3\times10^{-19}$ \\
AMOS left kidney & $86.01 \pm 4.8$ & $4.24 \pm 2.0$ & $4.14 \pm 2.1$ & $2.41 \pm 1.6$ & $\underline{8.57 \pm 15.7}$ & $0.87 \pm 3.2$ & $\mathbf{76.23 \pm 11.4}$ & $2.2\times10^{-29}$ \\
AMOS right kidney & $80.49 \pm 18.0$ & $3.80 \pm 2.2$ & $3.26 \pm 2.6$ & $4.06 \pm 1.7$ & $\underline{12.04 \pm 20.9}$ & -- & $\mathbf{59.68 \pm 24.2}$ & $1.7\times10^{-16}$ \\
AMOS spleen & $82.63 \pm 14.9$ & $1.34 \pm 2.2$ & $5.05 \pm 4.0$ & $4.53 \pm 2.7$ & $\underline{13.08 \pm 21.5}$ & $0.43 \pm 1.6$ & $\mathbf{69.00 \pm 20.6}$ & $1.9\times10^{-17}$ \\
AMOS stomach & $60.91 \pm 26.3$ & $4.50 \pm 4.8$ & $\underline{5.09 \pm 5.1}$ & $3.48 \pm 3.0$ & $4.89 \pm 9.5$ & -- & $\mathbf{49.88 \pm 26.4}$ & $1.3\times10^{-13}$ \\
\hline
ACDC RV cavity & $74.48 \pm 13.8$ & $0.65 \pm 0.5$ & $0.63 \pm 0.3$ & $0.11 \pm 0.2$ & $\underline{3.70 \pm 10.7}$ & -- & $\mathbf{57.87 \pm 19.4}$ & $1.1\times10^{-10}$ \\
ACDC LV cavity & $84.98 \pm 9.5$ & $1.02 \pm 1.0$ & $0.77 \pm 0.4$ & $1.01 \pm 0.6$ & $\underline{16.31 \pm 19.7}$ & -- & $\mathbf{77.82 \pm 11.9}$ & $1.0\times10^{-11}$ \\
ACDC myocardium & $76.21 \pm 4.6$ & $0.72 \pm 0.3$ & $3.69 \pm 0.9$ & $1.63 \pm 0.7$ & $\underline{25.07 \pm 16.2}$ & $0.66 \pm 3.0$ & $\mathbf{51.95 \pm 19.1}$ & $1.7\times10^{-6}$ \\
\hline
CAMUS Left Atrium & $91.54 \pm 4.5$ & $5.53 \pm 5.0$ & $1.23 \pm 1.3$ & $4.37 \pm 1.6$ & $\underline{16.61 \pm 4.5}$ & -- & $\mathbf{81.43 \pm 5.1}$ & $8.1\times10^{-26}$ \\
\hline
MCC left lung & $96.01 \pm 0.9$ & $16.25 \pm 10.8$ & $\underline{16.47 \pm 10.4}$ & $14.31 \pm 3.7$ & $3.00 \pm 4.0$ & -- & $\mathbf{92.13 \pm 2.0}$ & $7.5\times10^{-43}$ \\
MCC right lung & $95.63 \pm 1.0$ & $9.23 \pm 7.7$ & $4.22 \pm 1.5$ & $\underline{16.31 \pm 4.1}$ & $4.46 \pm 8.5$ & -- & $\mathbf{90.32 \pm 2.8}$ & $1.8\times10^{-32}$ \\
\hline
PMC polyp & $68.55 \pm 29.0$ & $11.49 \pm 16.5$ & $6.83 \pm 12.2$ & $10.25 \pm 8.5$ & $\underline{17.53 \pm 21.5}$ & -- & $\mathbf{78.02 \pm 17.8}$ & $2.5\times10^{-8}$ \\
\hline
\textbf{Average} & ${\color{gray}81.69 \pm 10.5}$ & $6.55 \pm 6.4$ & $5.77 \pm 5.7$ & $6.17 \pm 5.5$ & $\underline{11.70 \pm 6.9}$ & $0.17 \pm 0.3$ & $\mathbf{71.34 \pm 14.0}$ & $1.8\times10^{-158}$ \\
\hline
\end{tabular}
}
\end{table*}

\subsubsection{Evaluation Metrics}

We evaluate segmentation performance using the Dice Similarity Coefficient (DSC)~\cite{dice1945measures} and report target-wise DSC to compare performance across different medical targets.
For target confirmation, we report two interaction metrics, Grounding State Accuracy (GSA) and Target Confirmation Accuracy (TCA), on an evaluation set $\{(I_i, O_i, \theta_i^{\text{gt}})\}_{i=1}^{M}$ of $M$ instructions. Here, $I_i$ is a user instruction, $O_i$ the candidate set extracted by the Detection Agent, and $\theta_i^{\text{gt}}$ the ground-truth user-intended target.
GSA evaluates whether the Confirmation Agent correctly determines the grounding state of an instruction, i.e., whether the instruction is directly groundable or calls for clarification. With the grounding reliability threshold $\tau_1$, the predicted grounding state is $\hat{y}_i = \mathbf{1}[\,s_i \geq \tau_1\,]$ and the ground-truth grounding state is $y_i = \mathbf{1}[\,I_i \text{ is unambiguous}\,]$, where $\mathbf{1}[\cdot]$ is the indicator function and $s_i$ is the reliability score in~\eqref{eq:confidence}. GSA is the agreement between them:
\begin{equation}
    \mathrm{GSA} = \frac{1}{M}\sum_{i=1}^{M}\mathbf{1}\!\left[\,\hat{y}_i = y_i\,\right].
    \label{eq:gsa}
\end{equation}
TCA evaluates whether the final user-intended target $\theta_i$ in~\eqref{eq:confirm} matches the user-intended target $\theta_i^{\text{gt}}$. For directly grounded instructions ($s_i \geq \tau_1$), $\theta_i = \theta_i^{*}$. For ambiguous instructions ($s_i < \tau_1$), $\theta_i = \theta_i^{r}$ is the target confirmed after the clarification response $r$. TCA is defined as:
\begin{equation}
    \mathrm{TCA} = \frac{1}{M}\sum_{i=1}^{M}\mathbf{1}\!\left[\,\theta_i = \theta_i^{\text{gt}}\,\right],
    \label{eq:tca}
\end{equation}
where $\theta_i = \theta_i^{\text{gt}}$ holds when the confirmed candidate is the same target instance, in both category and location, as the intended one. The per-type TCA in Table~\ref{tab:target-confirmation} restricts this average to the corresponding instruction subset.

\subsection{Results and Analysis}

\begin{figure*}[t]
\centering
\includegraphics[width=\textwidth]{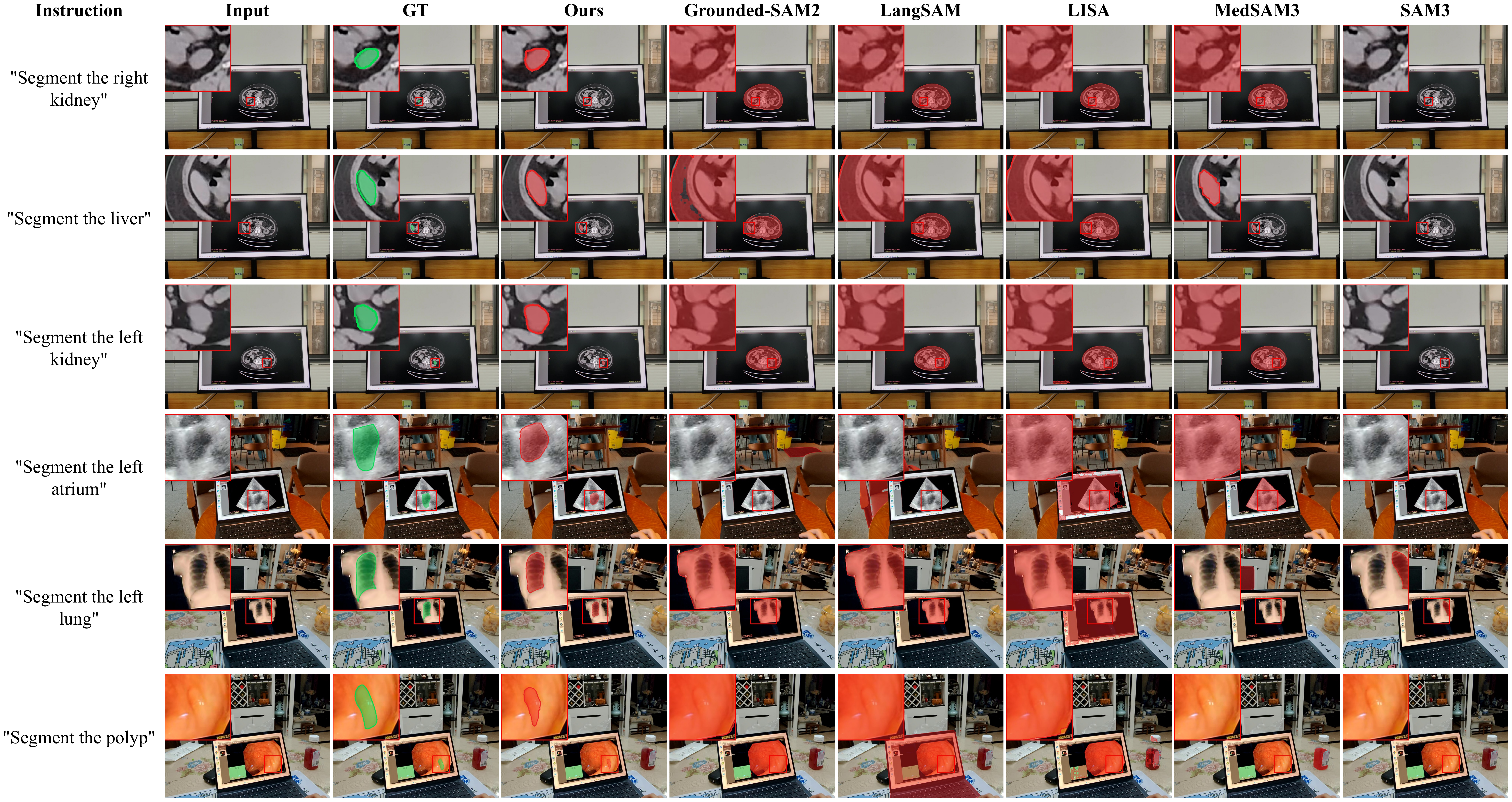}
\caption{On representative targets from different imaging scenarios, EgoMed-Agent produces masks concentrated on the user-intended target and close to the ground truth, while the text-prompted baselines produce over-expanded masks or segment irrelevant screen and background regions.}
\label{fig:qualitative}
\end{figure*}

\subsubsection{Quantitative Evaluation}

Table~\ref{tab:main_results} presents the quantitative comparison, from which we draw two observations.

\textit{a) EgoMed-Agent provides the most reliable instruction-guided segmentation.} It attains the highest Dice on all twelve targets among instruction-guided methods, with an average of $71.34\%$, and approaches the fully supervised nnU-Net upper bound ($81.69\%$) without training the segmentation model on mask annotations. This advantage comes from confirming the user-intended target before segmentation, so the mask is conditioned on an explicit target.

\textit{b) Directly grounding free-form instructions to masks is unreliable in IEMIS.} The text-prompted baselines remain below $12\%$ average Dice, and SAM3 fails to produce valid masks on most targets ($0.17\%$ average), indicating that open-vocabulary text grounding does not transfer to anatomical targets under egocentric visual conditions. Lacking an explicit confirmation step, these methods ground the instruction to screen or background regions, missing the intended object.

\subsubsection{Qualitative Analysis}

Fig.~\ref{fig:qualitative} shows qualitative segmentation results on representative targets from different imaging scenarios. Text-prompted segmentation methods ground the instruction unreliably in egocentric medical scenes. They produce over-expanded masks that cover a large part of the displayed medical image or the computer screen region, not the user-intended target. This is visible in targets such as the liver, kidneys, lung, and polyp, where the predicted masks from text-prompted baselines include non-target regions. In addition, for fine-grained anatomical structures such as kidneys and cardiac regions, these methods struggle to distinguish the target from nearby structures or visually similar regions.

In contrast, EgoMed-Agent produces masks that are more concentrated on the target regions and better aligned with the ground truth across different anatomical structures and lesions. This is because EgoMed-Agent first confirms the user-intended target through Target Confirmation and then performs Localization-Guided Propagation for the user-intended target, so it does not map the text instruction directly to a pixel-level mask. By separating target confirmation from mask generation, EgoMed-Agent reduces the risk of segmenting screen regions, background areas, or unintended neighboring structures, leading to more accurate segmentation results in egocentric medical scenes.

\subsubsection{Ablation of Components}
To isolate the roles of temporal propagation and localization guidance, we compare two reduced variants of EgoMed-Agent in Table~\ref{tab:module_ablation}. \textit{Init-Only Prop.} keeps only the Propagation Agent: it initializes from the first detector box and then propagates, which preserves temporal continuity but accumulates drift once the mask deviates from the user-intended target, so its average Dice drops to $50.46\%$. \textit{Frame-wise Det.} keeps only per-frame detection: it segments each frame independently, which avoids long-term drift but loses temporal continuity and remains sensitive to detection quality, reaching $59.27\%$. EgoMed-Agent couples the two, propagating for continuity while using detector-based localization to re-initialize when inconsistency occurs, and reaches $71.34\%$. This shows that temporal propagation and localization guidance are complementary.

\begin{table}[t]
\centering
\caption{Per-target Dice (\%) of EgoMed-Agent against its reduced variants Init-Only Prop. (propagation only) and Frame-wise Det. (per-frame detection only), where coupling propagation with detector-based localization gives the best Dice on every target.}
\label{tab:module_ablation}
\renewcommand{\arraystretch}{1.0}
\resizebox{\columnwidth}{!}{
\begin{tabular}{l|cc|c}
\thickhline
\textbf{Targets} & \textbf{Init-Only Prop.} & \textbf{Frame-wise Det.} & \textbf{EgoMed-Agent} \\
\hline
AMOS liver        & $39.86 \pm 32.6$ & $58.83 \pm 15.4$ & $\mathbf{71.69 \pm 15.6}$ \\
AMOS left kidney  & $63.00 \pm 28.0$ & $62.95 \pm 15.6$ & $\mathbf{76.23 \pm 11.4}$ \\
AMOS right kidney & $52.33 \pm 29.6$ & $43.07 \pm 23.5$ & $\mathbf{59.68 \pm 24.2}$ \\
AMOS spleen       & $40.98 \pm 30.1$ & $54.05 \pm 19.6$ & $\mathbf{69.00 \pm 20.6}$ \\
AMOS stomach      & $29.35 \pm 26.6$ & $39.46 \pm 22.7$ & $\mathbf{49.88 \pm 26.4}$ \\
\hline
ACDC RV cavity    & $52.13 \pm 21.8$ & $37.39 \pm 21.6$ & $\mathbf{57.87 \pm 19.4}$ \\
ACDC LV cavity    & $68.47 \pm 13.1$ & $66.11 \pm 14.1$ & $\mathbf{77.82 \pm 11.9}$ \\
ACDC myocardium   & $30.50 \pm 20.5$ & $33.22 \pm 17.7$ & $\mathbf{51.95 \pm 19.1}$ \\
\hline
CAMUS Left Atrium & $30.92 \pm 31.7$ & $74.27 \pm 11.6$ & $\mathbf{81.43 \pm 5.1}$  \\
\hline
MCC left lung     & $80.72 \pm 27.3$ & $88.96 \pm 3.2$  & $\mathbf{92.13 \pm 2.0}$  \\
MCC right lung    & $83.86 \pm 18.8$ & $85.49 \pm 6.4$  & $\mathbf{90.32 \pm 2.8}$  \\
\hline
PMC polyp         & $33.43 \pm 35.5$ & $67.45 \pm 23.8$ & $\mathbf{78.02 \pm 17.8}$ \\
\hline
\textbf{Average}  & $50.46 \pm 19.6$ & $59.27 \pm 18.5$ & $\mathbf{71.34 \pm 14.0}$ \\
\hline
\end{tabular}
}
\end{table}

\subsubsection{Ablation of Hyperparameters}
\label{ct}
We further analyze the effect of the consistency threshold $\tau_2$ in Localization-Guided Propagation. This threshold controls when localization-based re-initialization is triggered. When $\tau_2=0$, the system does not perform consistency-based re-initialization and mainly relies on continuous propagation by the Propagation Agent. As $\tau_2$ increases, the system becomes more sensitive to the inconsistency between the propagated result and the localization result, and therefore triggers re-initialization on more frames.

\begin{figure*}[t]
\centering
\includegraphics[width=\textwidth]{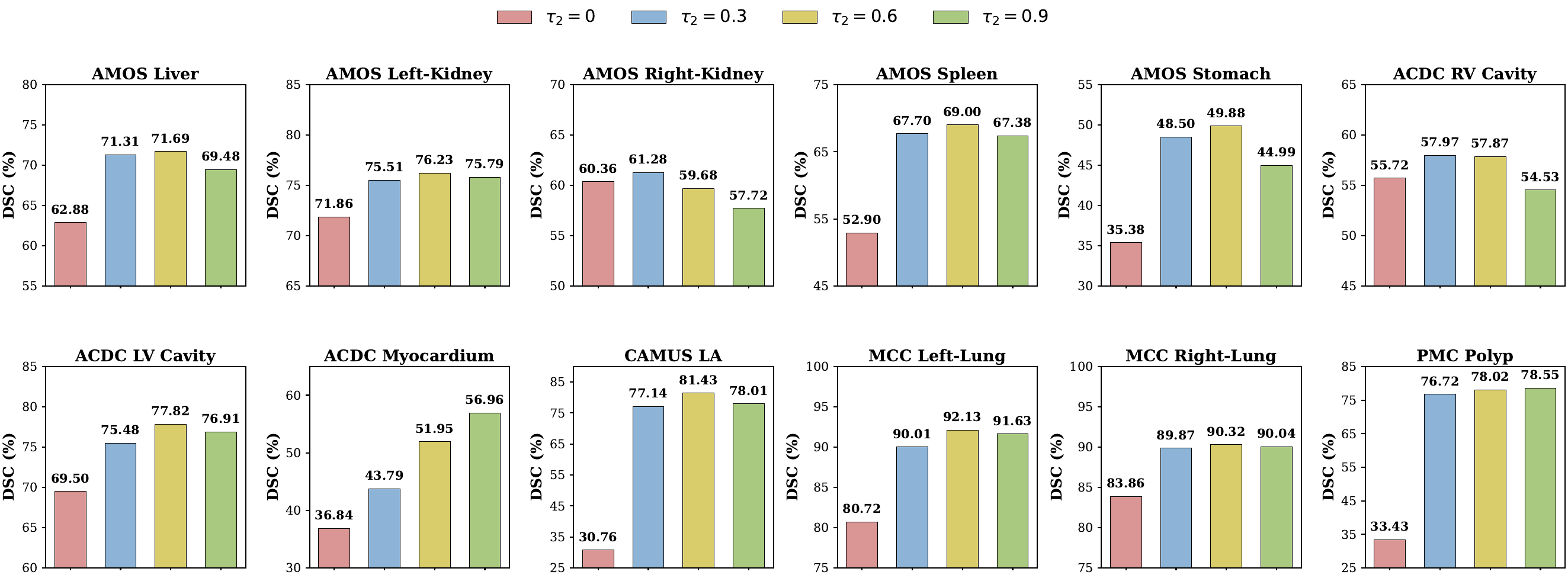}
\caption{A positive consistency threshold $\tau_2$ improves segmentation over $\tau_2=0$, and $\tau_2=0.6$ gives the best average Dice. Each subplot reports the Dice score under different $\tau_2$ settings for one medical target.}
\label{fig:threshold}
\end{figure*}

As shown in Fig.~\ref{fig:threshold}, a positive consistency threshold improves the average Dice score over $\tau_2=0$, indicating that our consistency evaluation mechanism is important for accurate segmentation across the egocentric video. Specifically, $\tau_2=0$, $\tau_2=0.3$, $\tau_2=0.6$, and $\tau_2=0.9$ achieve average DSC scores of $56.18\%$, $69.61\%$, $71.34\%$, and $70.17\%$ across all twelve targets, respectively. Among them, $\tau_2=0.6$ achieves the highest average DSC. When the threshold further increases to $\tau_2=0.9$, the performance does not continue to improve, suggesting that overly frequent re-initialization does not necessarily further benefit segmentation. Thus, we adopt $\tau_2=0.6$ as the consistency threshold in all of our main experiments.

\subsubsection{Effectiveness of Consistency Evaluation}
\begin{figure*}[t]
\centering
\includegraphics[width=\textwidth]{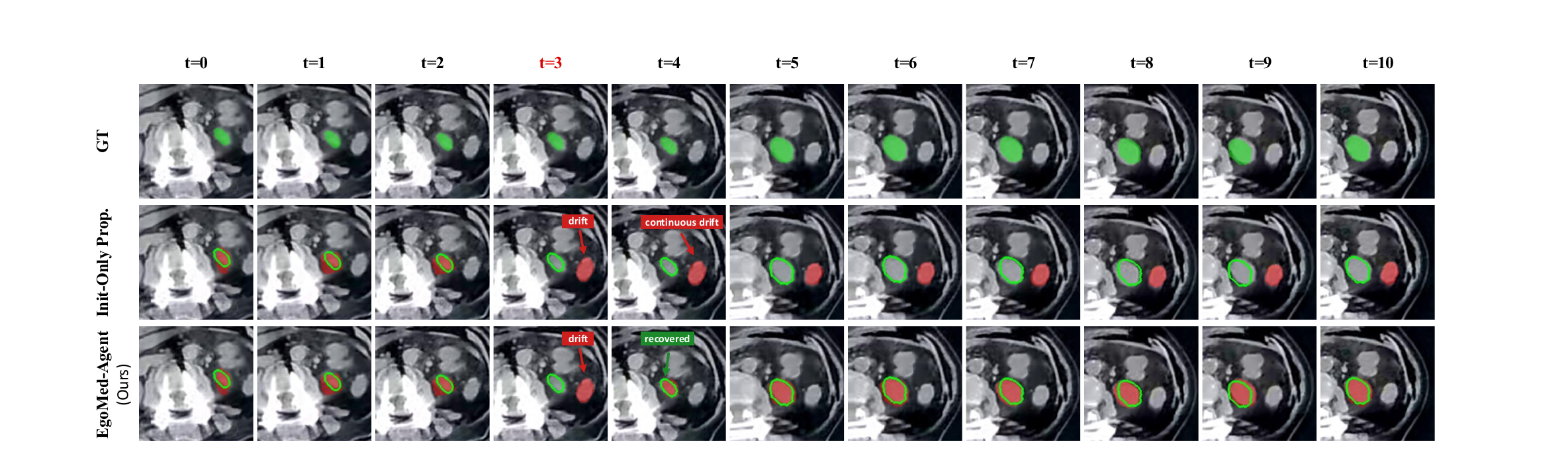}
\caption{Consistency Evaluation keeps EgoMed-Agent on the user-intended target where naive propagation drifts off it. The green contour denotes the ground-truth target and the red region denotes the predicted mask. From $t=0$ to $t=2$, both Init-Only Prop. and EgoMed-Agent follow the user-intended target. At $t=3$, Init-Only Prop. drifts toward a nearby non-target region and keeps propagating this erroneous mask, leaving its prediction off the target (the ``drift'' arrow). EgoMed-Agent detects the inconsistency and re-initializes from the localized target, so its prediction returns to the target (the ``recovered'' arrow).}
\label{fig:consistency}

\end{figure*}

Fig.~\ref{fig:consistency} illustrates the role of Consistency Evaluation during temporal propagation. From $t=0$ to $t=2$, both Init-Only Prop. and EgoMed-Agent follow the user-intended target. At $t=3$, Init-Only Prop. starts to drift toward a nearby non-target region. Relying only on continuous propagation, it carries this error forward to subsequent frames. In contrast, EgoMed-Agent detects the inconsistency between the propagated mask and the target-aware localization, and re-initializes propagation from the localized target position, correcting the drift and keeping the propagated mask aligned with the user-intended target. This correction is exercised throughout propagation: across the test set ($533$ video--target instances), EgoMed-Agent triggers on average $27.5$ re-initializations per target, $79.0\%$ of the targets trigger at least one, and the mean detection--propagation IoU is $0.787$. Together, these results show that drift is a recurring failure mode in egocentric medical video, and that Consistency Evaluation prevents error accumulation and improves segmentation accuracy.

\subsubsection{Inference Efficiency}
Table~\ref{tab:efficiency} reports the parameters and per-frame latency of EgoMed-Agent on a single NVIDIA RTX A6000. The full system has $113.0$M parameters and runs at $10.7$ FPS ($48.2$\,ms detection and $45.2$\,ms propagation per frame), using $3.7$\,GB of GPU memory. The Confirmation Agent (DeepSeek-V4-Flash) is queried once per instruction, not per frame, and is therefore excluded from the per-frame latency. EgoMed-Agent therefore runs in near-real-time on a single GPU, supporting interactive copilot assistance.

\begin{table}[t]
\centering
\caption{Inference efficiency of EgoMed-Agent on a single NVIDIA RTX A6000, with per-frame latency measured on a $200$-frame egocentric video.}
\label{tab:efficiency}
\renewcommand{\arraystretch}{1.0}
\resizebox{\columnwidth}{!}{
\begin{tabular}{lcc}
\thickhline
Component & Params (M) & Latency (ms/frame) \\
\hline
Detection Agent (YOLO26) & 21.8 & 48.2 \\
Propagation Agent (SAM2.1 b+) & 80.9 & 45.2 \\
Image-type classifier (YOLO26-cls) & 10.4 & once/video \\
\hline
\textbf{Total} & \textbf{113.0} & \textbf{93.4 (10.7\,FPS)} \\
\hline
\end{tabular}
}
\end{table}

\subsubsection{Target Confirmation Analysis}

We validate the Confirmation Agent on a 200-sample evaluation set balanced across four instruction types (unique semantic, unique spatial, ambiguous semantic, and ambiguous spatial), 50 samples each. Semantic instructions refer to the target by organ category (e.g., ``segment the liver''), while spatial instructions refer to it by an organ-relative relation. A relation that singles out one candidate is unique (e.g., ``the organ immediately to the left of the liver''), whereas one satisfied by multiple candidates is ambiguous (e.g., ``the organ on the left of the liver'' when several organs lie to its left) and calls for clarification.

As reported in Table~\ref{tab:target-confirmation}, with DeepSeek-V4-Flash the Confirmation Agent attains $100\%$ GSA and $100\%$ TCA, judging whether an instruction is directly groundable and, when ambiguous, determining the user-intended target through clarification. On the same set, a lighter LLM (DeepSeek-Chat) reaches only $80\%$ GSA and $70.5\%$ TCA, with errors concentrated on organ-relative spatial instructions ($44\%$ and $38\%$ TCA on unique and ambiguous spatial). This indicates that spatial grounding is the main source of error and benefits from a stronger reasoner. Overall, these results show that Target Confirmation evaluates grounding reliability and either confirms the target or requests clarification, so the subsequent segmentation is conditioned on an explicit target.

\begin{table}[t]
\centering
\caption{Per-type target confirmation accuracy (TCA, \%) of the EgoMed-Agent Confirmation Agent on the 200-sample set under two LLM backends, where each instruction type contains 50 samples. Over all samples, GSA\,/\,TCA reach 80\,/\,70.5 with DS-Chat and 100\,/\,100 with DS-V4Flash, showing that the stronger backend removes the spatial-grounding errors of the lighter one.}
\label{tab:target-confirmation}
\renewcommand{\arraystretch}{1.0}
\resizebox{\columnwidth}{!}{
\begin{tabular}{lccc}
\thickhline
Instruction Type & Num. & DS-Chat & DS-V4Flash \\
\hline
Unique semantic & 50 & 100 & 100 \\
Unique spatial & 50 & 44 & 100 \\
\hline
Ambiguous semantic & 50 & 100 & 100 \\
Ambiguous spatial & 50 & 38 & 100 \\
\hline
\end{tabular}
}
\end{table}

\section{Discussion and Conclusion}

This paper studied IEMIS, the problem of segmenting a user-intended target from the egocentric video of medical smart glasses. We proposed EgoMed-Agent, a multi-agent system in which a Detection Agent, a Confirmation Agent, and a Propagation Agent cooperate with the user to confirm the user-intended target and keep the segmentation locked onto it as the egocentric view changes. On our IEMIS benchmark, EgoMed-Agent reached $71.34\%$ average Dice and significantly outperformed the strongest text-prompted baseline. By understanding the target from the human perspective, EgoMed-Agent offers a general paradigm for reliable instruction-guided medical perception on smart-glasses copilots.

Two aspects remain open and define our future work. First, EgoMed-Agent relies on its detection toolbox for candidate targets and localization, which bounds the covered modalities, anatomical structures, and lesion types. This is alleviated by extending the toolbox with additional specialized detectors and open-vocabulary detection~\cite{liu2023grounding}, which we will pursue to broaden coverage. Second, our dataset is built from screen-displayed public medical images captured by smart glasses. Although this reflects a practical smart-glasses-assisted scenario, it does not yet cover all real clinical deployment conditions, and collecting egocentric data directly from clinical imaging workstations is a further direction of our future work.

\section*{REFERENCES}
\vspace{-2em}
\bibliographystyle{IEEEtran}
\bibliography{references}

\begin{thebibliography}{10}
\providecommand{\url}[1]{#1}
\csname url@samestyle\endcsname
\providecommand{\newblock}{\relax}
\providecommand{\bibinfo}[2]{#2}
\providecommand{\BIBentrySTDinterwordspacing}{\spaceskip=0pt\relax}
\providecommand{\BIBentryALTinterwordstretchfactor}{4}
\providecommand{\BIBentryALTinterwordspacing}{\spaceskip=\fontdimen2\font plus
\BIBentryALTinterwordstretchfactor\fontdimen3\font minus
  \fontdimen4\font\relax}
\providecommand{\BIBforeignlanguage}[2]{{%
\expandafter\ifx\csname l@#1\endcsname\relax
\typeout{** WARNING: IEEEtran.bst: No hyphenation pattern has been}%
\typeout{** loaded for the language `#1'. Using the pattern for}%
\typeout{** the default language instead.}%
\else
\language=\csname l@#1\endcsname
\fi
#2}}
\providecommand{\BIBdecl}{\relax}
\BIBdecl

\bibitem{malungana2025use}
L.~Malungana and B.~Chimbo, ``The use of smart glasses in the healthcare
  metaverse: A systematic review,'' \emph{Suid-Afrikaanse Tydskrif vir
  Natuurwetenskap en Tegnologie}, vol.~44, no.~1, pp. 8--16, 2025.

\bibitem{zhang2022designing}
Z.~Zhang, N.~A. Ramiya Ramesh~Babu, K.~Adelgais, and M.~Ozkaynak, ``Designing
  and implementing smart glass technology for emergency medical services: a
  sociotechnical perspective,'' \emph{JAMIA open}, vol.~5, no.~4, p. ooac113,
  2022.

\bibitem{waite2017interpretive}
S.~Waite, J.~Scott, B.~Gale, T.~Fuchs, S.~Kolla, and D.~Reede, ``Interpretive
  error in radiology,'' \emph{American Journal of Roentgenology}, vol. 208,
  no.~4, pp. 739--749, 2017.

\bibitem{yu2014radiologist}
J.-P.~J. Yu, A.~P. Kansagra, and J.~Mongan, ``The radiologist's workflow
  environment: evaluation of disruptors and potential implications,''
  \emph{Journal of the American College of Radiology}, vol.~11, no.~6, pp.
  589--593, 2014.

\bibitem{drew2013invisible}
T.~Drew, M.~L.-H. V{\~o}, and J.~M. Wolfe, ``The invisible gorilla strikes
  again: sustained inattentional blindness in expert observers,''
  \emph{Psychological Science}, vol.~24, no.~9, pp. 1848--1853, 2013.

\bibitem{su2025ova}
H.~Su, M.~Xie, N.~Cao, Y.~Ding, B.~Shao, X.~Long, F.~Gu, and C.~Chen,
  ``Ova-fields: Weakly supervised open-vocabulary affordance fields for robot
  operational part detection,'' in \emph{Proceedings of the IEEE/CVF
  International Conference on Computer Vision}, 2025, pp. 6385--6395.

\bibitem{zhang2024clamber}
T.~Zhang, P.~Qin, Y.~Deng, C.~Huang, W.~Lei, J.~Liu, D.~Jin, H.~Liang, and
  T.-S. Chua, ``Clamber: A benchmark of identifying and clarifying ambiguous
  information needs in large language models,'' in \emph{Proceedings of the
  62nd Annual Meeting of the Association for Computational Linguistics (Volume
  1: Long Papers)}, 2024, pp. 10\,746--10\,766.

\bibitem{ding2026open}
J.~Ding, H.~Tang, and G.~Li, ``Open-vocabulary 3d instruction ambiguity
  detection,'' \emph{arXiv preprint arXiv:2601.05991}, 2026.

\bibitem{li2026challenges}
X.~Li, H.~Qiu, L.~Wang, H.~Zhang, C.~Qi, L.~Han, H.~Xiong, and H.~Li,
  ``Challenges and trends in egocentric vision: A survey,'' \emph{Machine
  Intelligence Research}, vol.~23, no.~1, pp. 1--33, 2026.

\bibitem{poleg2014temporal}
Y.~Poleg, C.~Arora, and S.~Peleg, ``Temporal segmentation of egocentric
  videos,'' in \emph{Proceedings of the IEEE conference on computer vision and
  pattern recognition}, 2014, pp. 2537--2544.

\bibitem{millerdurai2024eventego3d}
C.~Millerdurai, H.~Akada, J.~Wang, D.~Luvizon, C.~Theobalt, and V.~Golyanik,
  ``Eventego3d: 3d human motion capture from egocentric event streams,'' in
  \emph{Proceedings of the IEEE/CVF Conference on Computer Vision and Pattern
  Recognition}, 2024, pp. 1186--1195.

\bibitem{wang2025caldiff}
X.~Wang, M.~Yang, S.~Tosun, K.~Nakamura, S.~Li, and X.~Li, ``Caldiff:
  calibrating uncertainty and accessing reliability of diffusion models for
  trustworthy lesion segmentation,'' \emph{IEEE journal of biomedical and
  health informatics}, 2025.

\bibitem{li2024agent}
A.~Li, Y.~Xie, S.~Li, F.~Tsung, B.~Ding, and Y.~Li, ``Agent-oriented planning
  in multi-agent systems,'' \emph{arXiv preprint arXiv:2410.02189}, 2024.

\bibitem{zhang2025agentorchestra}
W.~Zhang, C.~Cui, Y.~Zhao, R.~Hu, Y.~Liu, Y.~Zhou, and B.~An, ``Agentorchestra:
  A hierarchical multi-agent framework for general-purpose task solving,''
  \emph{arXiv e-prints}, pp. arXiv--2506, 2025.

\bibitem{hu2026evolutionary}
Y.~Hu, M.~Trager, Y.~Zhang, Y.~Zhang, S.~Yang, W.~Xia, and S.~Soatto,
  ``Evolutionary generation of multi-agent systems,'' \emph{arXiv preprint
  arXiv:2602.06511}, 2026.

\bibitem{damen2018scaling}
D.~Damen, H.~Doughty, G.~M. Farinella, S.~Fidler, A.~Furnari, E.~Kazakos,
  D.~Moltisanti, J.~Munro, T.~Perrett, W.~Price \emph{et~al.}, ``Scaling
  egocentric vision: The epic-kitchens dataset,'' in \emph{Proceedings of the
  European conference on computer vision (ECCV)}, 2018, pp. 720--736.

\bibitem{grauman2022ego4d}
K.~Grauman, A.~Westbury, E.~Byrne, Z.~Chavis, A.~Furnari, R.~Girdhar,
  J.~Hamburger, H.~Jiang, M.~Liu, X.~Liu \emph{et~al.}, ``Ego4d: Around the
  world in 3,000 hours of egocentric video,'' in \emph{Proceedings of the
  IEEE/CVF conference on computer vision and pattern recognition}, 2022, pp.
  18\,995--19\,012.

\bibitem{plizzari2025omnia}
C.~Plizzari, A.~Tonioni, Y.~Xian, A.~Kulshrestha, and F.~Tombari, ``Omnia de
  egotempo: Benchmarking temporal understanding of multi-modal llms in
  egocentric videos,'' in \emph{Proceedings of the Computer Vision and Pattern
  Recognition Conference}, 2025, pp. 24\,129--24\,138.

\bibitem{zhou2025egotextvqa}
S.~Zhou, J.~Xiao, Q.~Li, Y.~Li, X.~Yang, D.~Guo, M.~Wang, T.-S. Chua, and
  A.~Yao, ``Egotextvqa: Towards egocentric scene-text aware video question
  answering,'' in \emph{Proceedings of the Computer Vision and Pattern
  Recognition Conference}, 2025, pp. 3363--3373.

\bibitem{liu2025ceres}
H.~Liu, Z.~Song, H.~Wu, T.~Pu, K.~Wang, and L.~Lin, ``Robust egocentric
  referring video object segmentation via dual-modal causal intervention,'' in
  \emph{Advances in Neural Information Processing Systems}, 2025.

\bibitem{shi2024cognition}
Z.~Shi, H.~Qiu, L.~Wang, F.~Meng, Q.~Wu, and H.~Li, ``Cognition transferring
  and decoupling for text-supervised egocentric semantic segmentation,''
  \emph{arXiv preprint arXiv:2410.01341}, 2024.

\bibitem{fujii2024egosurgery}
R.~Fujii, M.~Hatano, H.~Saito, and H.~Kajita, ``Egosurgery-phase: A dataset of
  surgical phase recognition from egocentric open surgery videos,'' in
  \emph{International Conference on Medical Image Computing and
  Computer-Assisted Intervention}.\hskip 1em plus 0.5em minus 0.4em\relax
  Springer, 2024, pp. 187--196.

\bibitem{zhuo2025egocentric}
Y.~Zhuo, E.~Zhang, X.~Yu, A.~Pachpande, W.~Fang, X.~Chen, A.~W. Kirkpatrick,
  K.~Couperus, C.~Colombo, O.~Tran \emph{et~al.}, ``An egocentric life-saving
  interventional procedure dataset of actions, medical questions, maneuvers and
  tools,'' \emph{Scientific Data}, 2025.

\bibitem{darjana2025egosurgery}
N.~Darjana, R.~Fujii, H.~Saito, and H.~Kajita, ``Egosurgery-hts: A dataset for
  egocentric hand--tool segmentation in open surgery videos,'' \emph{Healthcare
  Technology Letters}, vol.~12, no.~1, p. e70049, 2025.

\bibitem{plizzari2024outlook}
C.~Plizzari, G.~Goletto, A.~Furnari, S.~Bansal, F.~Ragusa, G.~M. Farinella,
  D.~Damen, and T.~Tommasi, ``An outlook into the future of egocentric vision:
  C. plizzari et al.'' \emph{International Journal of Computer Vision}, vol.
  132, no.~11, pp. 4880--4936, 2024.

\bibitem{yang2026egocentric}
S.~Yang, Y.~Huang, W.~Cai, S.~Sun, F.~Fang, Y.~He, Y.~Xie, J.~Deng, H.~Zhang,
  J.~Song, and Z.~Zhang, ``Egocentric co-pilot: Web-native smart-glasses agents
  for assistive egocentric ai,'' \emph{arXiv preprint arXiv:2603.01104}, 2026.

\bibitem{gollapalli2024smart}
S.~Gollapalli, V.~Sharma, A.~Al~Ghazwi, and L.~Heskin, ``Smart glasses in
  surgery: the theatre and beyond,'' \emph{Surgical Innovation}, vol.~31,
  no.~5, pp. 502--508, 2024.

\bibitem{mitrasinovic2015clinical}
S.~Mitrasinovic, E.~Camacho, N.~Trivedi, J.~Logan, C.~Campbell, R.~Zilinyi,
  B.~Lieber, E.~Bruce, B.~Taylor, D.~Martineau \emph{et~al.}, ``Clinical and
  surgical applications of smart glasses,'' \emph{Technology and health care},
  vol.~23, no.~4, pp. 381--401, 2015.

\bibitem{lu2024multimodal}
M.~Y. Lu, B.~Chen, D.~F. Williamson, R.~J. Chen, M.~Zhao, A.~K. Chow,
  K.~Ikemura, A.~Kim, D.~Pouli, A.~Patel \emph{et~al.}, ``A multimodal
  generative ai copilot for human pathology,'' \emph{Nature}, vol. 634, no.
  8033, pp. 466--473, 2024.

\bibitem{xu2024advances}
Y.~Xu, R.~Quan, W.~Xu, Y.~Huang, X.~Chen, and F.~Liu, ``Advances in medical
  image segmentation: A comprehensive review of traditional, deep learning and
  hybrid approaches,'' \emph{Bioengineering}, vol.~11, no.~10, p. 1034, 2024.

\bibitem{ma2024segment}
J.~Ma, Y.~He, F.~Li, L.~Han, C.~You, and B.~Wang, ``Segment anything in medical
  images,'' \emph{Nature communications}, vol.~15, no.~1, p. 654, 2024.

\bibitem{wang2018deepigeos}
G.~Wang, M.~A. Zuluaga, W.~Li, R.~Pratt, P.~A. Patel, M.~Aertsen, T.~Doel,
  A.~L. David, J.~Deprest, S.~Ourselin \emph{et~al.}, ``Deepigeos: a deep
  interactive geodesic framework for medical image segmentation,'' \emph{IEEE
  transactions on pattern analysis and machine intelligence}, vol.~41, no.~7,
  pp. 1559--1572, 2018.

\bibitem{heinemann2025limis}
L.~Heinemann, A.~Jaus, Z.~Marinov, M.~Kim, M.~F. Spadea, J.~Kleesiek, and
  R.~Stiefelhagen, ``Limis: Towards language-based interactive medical image
  segmentation,'' in \emph{2025 IEEE 22nd International Symposium on Biomedical
  Imaging (ISBI)}.\hskip 1em plus 0.5em minus 0.4em\relax IEEE, 2025, pp. 1--5.

\bibitem{yuan2025tgsam}
R.~Yuan, L.~Zhou, J.~Xu, Q.~Li, M.~Chen, Y.~Zhang, R.~Feng, T.~Zhang, and
  S.~Gao, ``Tgsam-2: Text-guided medical image segmentation using segment
  anything model 2,'' in \emph{International Conference on Medical Image
  Computing and Computer-Assisted Intervention}.\hskip 1em plus 0.5em minus
  0.4em\relax Springer, 2025, pp. 565--574.

\bibitem{liu2025medsam3}
A.~Liu, R.~Xue, X.~R. Cao, Y.~Shen, Y.~Lu, X.~Li, Q.~Chen, and J.~Chen,
  ``Medsam3: Delving into segment anything with medical concepts,'' \emph{arXiv
  preprint arXiv:2511.19046}, 2025.

\bibitem{lai2024lisa}
X.~Lai, Z.~Tian, Y.~Chen, Y.~Li, Y.~Yuan, S.~Liu, and J.~Jia, ``Lisa: Reasoning
  segmentation via large language model,'' in \emph{Proceedings of the IEEE/CVF
  conference on computer vision and pattern recognition}, 2024, pp. 9579--9589.

\bibitem{li2024prism}
H.~Li, H.~Liu, D.~Hu, J.~Wang, and I.~Oguz, ``Prism: A promptable and robust
  interactive segmentation model with visual prompts,'' in \emph{International
  Conference on Medical Image Computing and Computer-Assisted
  Intervention}.\hskip 1em plus 0.5em minus 0.4em\relax Springer, 2024, pp.
  389--399.

\bibitem{xie2024simtxtseg}
Y.~Xie, T.~Zhou, Y.~Zhou, and G.~Chen, ``Simtxtseg: Weakly-supervised medical
  image segmentation with simple text cues,'' in \emph{International Conference
  on Medical Image Computing and Computer-Assisted Intervention}.\hskip 1em
  plus 0.5em minus 0.4em\relax Springer, 2024, pp. 634--644.

\bibitem{shen2023temporally}
C.~Shen, W.~Li, Y.~Zhang, Y.~Wang, and X.~Wang, ``Temporally-extended prompts
  optimization for sam in interactive medical image segmentation,'' in
  \emph{2023 IEEE International Conference on Bioinformatics and Biomedicine
  (BIBM)}.\hskip 1em plus 0.5em minus 0.4em\relax IEEE, 2023, pp. 3550--3557.

\bibitem{schmidgall2024agentclinic}
S.~Schmidgall, R.~Ziaei, C.~Harris, E.~Reis, J.~Jopling, and M.~Moor,
  ``Agentclinic: a multimodal agent benchmark to evaluate ai in simulated
  clinical environments,'' \emph{arXiv preprint arXiv:2405.07960}, 2024.

\bibitem{almansoori2025medagentsim}
M.~Almansoori, K.~Kumar, and H.~Cholakkal, ``Medagentsim: Self-evolving
  multi-agent simulations for realistic clinical interactions,'' in
  \emph{International Conference on Medical Image Computing and
  Computer-Assisted Intervention}.\hskip 1em plus 0.5em minus 0.4em\relax
  Springer, 2025, pp. 362--372.

\bibitem{elboardy2025medical}
A.~T. Elboardy, G.~Khoriba, and E.~A. Rashed, ``Medical ai consensus: A
  multi-agent framework for radiology report generation and evaluation,''
  \emph{arXiv preprint arXiv:2509.17353}, 2025.

\bibitem{yi2025multimodal}
Z.~Yi, T.~Xiao, and M.~V. Albert, ``A multimodal multi-agent framework for
  radiology report generation,'' \emph{arXiv preprint arXiv:2505.09787}, 2025.

\bibitem{liao2025reflectool}
Y.~Liao, S.~Jiang, Y.~Wang, and Y.~Wang, ``Reflectool: Towards reflection-aware
  tool-augmented clinical agents,'' in \emph{Proceedings of the 63rd Annual
  Meeting of the Association for Computational Linguistics (Volume 1: Long
  Papers)}, 2025, pp. 13\,507--13\,531.

\bibitem{ji2022amos}
Y.~Ji, H.~Bai, C.~Ge, J.~Yang, Y.~Zhu, R.~Zhang, Z.~Li, L.~Zhanng, W.~Ma,
  X.~Wan \emph{et~al.}, ``Amos: A large-scale abdominal multi-organ benchmark
  for versatile medical image segmentation,'' \emph{Advances in neural
  information processing systems}, vol.~35, pp. 36\,722--36\,732, 2022.

\bibitem{bernard2018deep}
O.~Bernard, A.~Lalande, C.~Zotti, F.~Cervenansky, X.~Yang, P.-A. Heng,
  I.~Cetin, K.~Lekadir, O.~Camara, M.~A.~G. Ballester \emph{et~al.}, ``Deep
  learning techniques for automatic mri cardiac multi-structures segmentation
  and diagnosis: is the problem solved?'' \emph{IEEE transactions on medical
  imaging}, vol.~37, no.~11, pp. 2514--2525, 2018.

\bibitem{leclerc2019deep}
S.~Leclerc, E.~Smistad, J.~Pedrosa, A.~{\O}stvik, F.~Cervenansky, F.~Espinosa,
  T.~Espeland, E.~A.~R. Berg, P.-M. Jodoin, T.~Grenier \emph{et~al.}, ``Deep
  learning for segmentation using an open large-scale dataset in 2d
  echocardiography,'' \emph{IEEE transactions on medical imaging}, vol.~38,
  no.~9, pp. 2198--2210, 2019.

\bibitem{jaeger2014two}
S.~Jaeger, S.~Candemir, S.~Antani, Y.-X.~J. W{\'a}ng, P.-X. Lu, and G.~Thoma,
  ``Two public chest x-ray datasets for computer-aided screening of pulmonary
  diseases,'' \emph{Quantitative imaging in medicine and surgery}, vol.~4,
  no.~6, p. 475, 2014.

\bibitem{ali2023multi}
S.~Ali, D.~Jha, N.~Ghatwary, S.~Realdon, R.~Cannizzaro, O.~E. Salem,
  D.~Lamarque, C.~Daul, M.~A. Riegler, K.~V. Anonsen \emph{et~al.}, ``A
  multi-centre polyp detection and segmentation dataset for generalisability
  assessment,'' \emph{Scientific Data}, vol.~10, no.~1, p.~75, 2023.

\bibitem{ravi2024sam2segmentimages}
\BIBentryALTinterwordspacing
N.~Ravi, V.~Gabeur, Y.-T. Hu, R.~Hu, C.~Ryali, T.~Ma, H.~Khedr, R.~Rädle,
  C.~Rolland, L.~Gustafson, E.~Mintun, J.~Pan, K.~V. Alwala, N.~Carion, C.-Y.
  Wu, R.~Girshick, P.~Dollár, and C.~Feichtenhofer, ``Sam 2: Segment anything
  in images and videos,'' 2024. [Online]. Available:
  \url{https://arxiv.org/abs/2408.00714}
\BIBentrySTDinterwordspacing

\bibitem{liu2023grounding}
S.~Liu, Z.~Zeng, T.~Ren, F.~Li, H.~Zhang, J.~Yang, C.~Li, J.~Yang, H.~Su,
  J.~Zhu \emph{et~al.}, ``Grounding dino: Marrying dino with grounded
  pre-training for open-set object detection,'' \emph{arXiv preprint
  arXiv:2303.05499}, 2023.

\bibitem{ren2024grounding}
T.~Ren, Q.~Jiang, S.~Liu, Z.~Zeng, W.~Liu, H.~Gao, H.~Huang, Z.~Ma, X.~Jiang,
  Y.~Chen, Y.~Xiong, H.~Zhang, F.~Li, P.~Tang, K.~Yu, and L.~Zhang, ``Grounding
  dino 1.5: Advance the "edge" of open-set object detection,'' 2024.

\bibitem{ren2024grounded}
T.~Ren, S.~Liu, A.~Zeng, J.~Lin, K.~Li, H.~Cao, J.~Chen, X.~Huang, Y.~Chen,
  F.~Yan, Z.~Zeng, H.~Zhang, F.~Li, J.~Yang, H.~Li, Q.~Jiang, and L.~Zhang,
  ``Grounded sam: Assembling open-world models for diverse visual tasks,''
  2024.

\bibitem{kirillov2023segany}
A.~Kirillov, E.~Mintun, N.~Ravi, H.~Mao, C.~Rolland, L.~Gustafson, T.~Xiao,
  S.~Whitehead, A.~C. Berg, W.-Y. Lo, P.~Doll{\'a}r, and R.~Girshick, ``Segment
  anything,'' \emph{arXiv:2304.02643}, 2023.

\bibitem{langsam}
L.~Medeiros and contributors, ``Language segment-anything,''
  \url{https://github.com/luca-medeiros/lang-segment-anything}, 2023, gitHub
  repository.

\bibitem{carion2025sam}
N.~Carion, L.~Gustafson, Y.-T. Hu, S.~Debnath, R.~Hu, D.~Suris, C.~Ryali, K.~V.
  Alwala, H.~Khedr, A.~Huang \emph{et~al.}, ``Sam 3: Segment anything with
  concepts,'' \emph{arXiv preprint arXiv:2511.16719}, 2025.

\bibitem{isensee2021nnu}
F.~Isensee, P.~F. Jaeger, S.~A. Kohl, J.~Petersen, and K.~H. Maier-Hein,
  ``nnu-net: a self-configuring method for deep learning-based biomedical image
  segmentation,'' \emph{Nature methods}, vol.~18, no.~2, pp. 203--211, 2021.

\bibitem{dice1945measures}
L.~R. Dice, ``Measures of the amount of ecologic association between species,''
  \emph{Ecology}, vol.~26, no.~3, pp. 297--302, 1945.

\end{thebibliography}
\end{document}